\documentclass[10pt,twocolumn,letterpaper]{article}
\PassOptionsToPackage{table}{xcolor}
\usepackage{wacv}              

\usepackage[table]{xcolor}
\usepackage{booktabs}
\usepackage{tikz}
\usepackage{microtype}
\usepackage[accsupp]{axessibility}

\usepackage{wasysym}
\usepackage{graphicx}

\definecolor{missorange}{RGB}{240,120,45}
\definecolor{hallred}{RGB}{190,0,0}

\newcommand{\thickwhitecross}{%
  \makebox[0.45em][c]{%
    \raisebox{0pt}[0pt][0pt]{%
      \ooalign{%
        \hfil\rotatebox[origin=c]{45}{\rule{0.6em}{0.12em}}\hfil\cr
        \hfil\rotatebox[origin=c]{-45}{\rule{0.6em}{0.12em}}\hfil\cr
      }%
    }%
  }%
}

\DeclareRobustCommand{\filledcross}[1]{%
  \raisebox{-0.12ex}{%
    \ooalign{%
      \hfil\textcolor{#1}{\Large\CIRCLE}\hfil\cr
      \hfil\raisebox{0.80ex}{\textcolor{white}{\thickwhitecross}}\hfil\cr
    }%
  }%
}

\DeclareRobustCommand{\orangecross}{\filledcross{missorange}}
\DeclareRobustCommand{\redcross}{\filledcross{hallred}}

\makeatletter
\def\wacvsection{\@startsection {section}{1}{\z@}
   {-8pt plus -2pt minus -2pt}{5pt} {\large\bf}}
\def\wacvsubsection{\@startsection {subsection}{2}{\z@}
   {-6pt plus -2pt minus -2pt}{4pt} {\elvbf}}
\def\wacvsubsubsection{\@startsection {subsubsection}{3}{\z@}
   {-5pt plus -1pt minus -1pt}{2pt} {\tenbf}}
\makeatother
\definecolor{wacvblue}{rgb}{0.21,0.49,0.74}
\usepackage[pagebackref,breaklinks,colorlinks,allcolors=wacvblue]{hyperref}

\def\wacvPaperID{492} 
\def\confName{WACV}
\def\confYear{2027}

\title{Why Domain Matters: Domain-Aware Benchmarking of\\Underwater Object Detection and Annotation Quality}

\author{
Melanie Wille \qquad
Dimity Miller \qquad
Tobias Fischer \qquad
Scarlett Raine\\
QUT Centre for Robotics, Queensland University of Technology\\
Brisbane, Australia\\
{\tt\small \{willemc,d24.miller,tobias.fischer,sg.raine\}@qut.edu.au}
}

\begin{document}
\maketitle
\begin{abstract}
Underwater object detection is strongly affected by domain shift, where performance can vary significantly across different locations, habitats, and deployment conditions. However, detector performance is typically evaluated using aggregate metrics that hide failures in specific environments, while existing domain generalization benchmarks often rely on synthetic variations that do not reflect real-world conditions. To address these limitations we introduce a framework that characterizes underwater images by appearance, scene composition, and acquisition geometry to assign domain labels. Using this framework, we perform the first systematic study of how domain factors influence both human annotation quality in underwater object detection datasets and deep learning-based detector performance, revealing substantial domain-dependent discrepancies. By incorporating physically meaningful domain labels, domain shift becomes something we can characterize, measure, benchmark, and act on. We highlight how our framework can be used to guide data collection and annotation, design more informative benchmarks, and assess detector robustness across diverse underwater environments.
\end{abstract}

\section{Introduction}
\label{sec:intro}

Underwater object detection is an important tool for marine science, enabling large-scale analysis of seafloor imagery to monitor benthic indicator species and assess ecosystem health under increasing human activity~\cite{doig2025training}. However, detector development often relies on collecting and annotating large amounts of training data, a process that is expensive and difficult to scale across new sites, habitats, and collection protocols~\cite{han2023see, walker2024domain}. A major driver of data requirements is the large variation of underwater conditions, such as lighting, turbidity, depth, and environmental structure, which alter both image appearance and object characteristics, creating a wide range of distinct domains~\cite{raine2026ai}. This often causes models trained on one set of conditions to perform poorly when deployed in another, manifesting as domain shift, which challenges model generalization and limits data reusability~\cite{han2023see, elmezain2025advancing, nabahirwa2025structured}.

\begin{figure}
    \centering
    \includegraphics[
        width=\linewidth,
        trim={0cm 0.3cm 0cm 0cm},
        clip
    ]{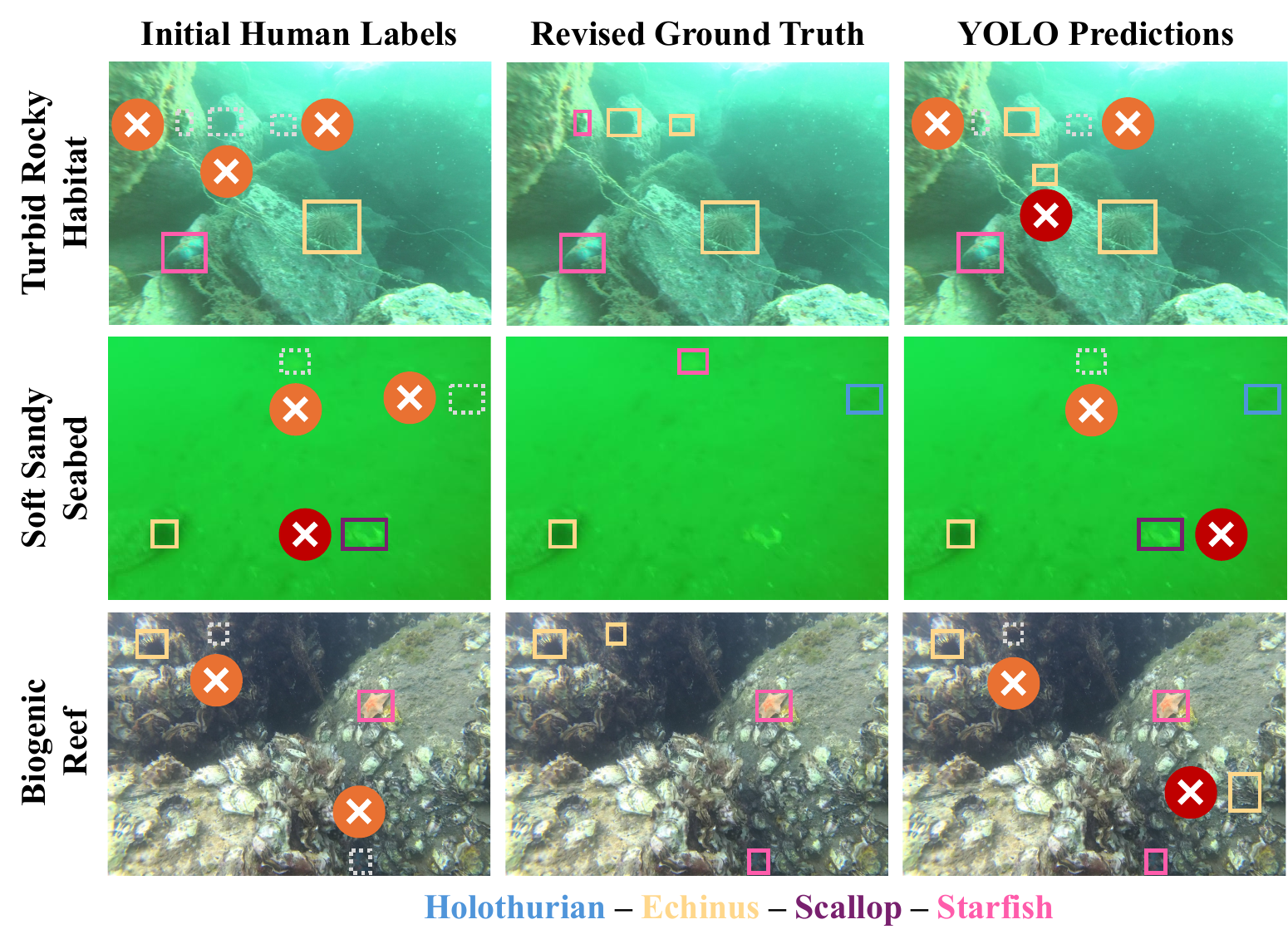}
    \vspace{-0.7cm}
    \caption{
    Example underwater images from visually distinct underwater domains. Left: original RUOD labels. Middle: revised RUOD-R labels (considered closest to reality). Right: YOLO26n predictions. Some true objects were missed (orange crosses\,\orangecross), while others were hallucinated (red crosses\,\redcross). Domain conditions challenge both humans and detectors. 
    }
    \label{fig:front_page}
    \vspace{-0.3cm}
\end{figure}

Fig.~\ref{fig:front_page} illustrates this challenge: under changing underwater conditions, the same target classes may differ substantially in appearance and therefore detectability. In this paper, we investigate how such domain differences influence the performance of both object detectors and human annotators for consistently identifying and localizing targets in an image. Understanding the human side of the problem is particularly important because annotation quality directly determines the reliability of training and evaluation data, yet is typically assumed to be uniform across conditions.

Existing underwater object detection benchmarks provide valuable resources for training and evaluating detectors~\cite{pedersen2019brackish, liu2021duo, fu2023ruod, jiang2021uodd}, but they are not designed to explain domain-dependent failures. In this work, we use the term \emph{domain} to refer to a subset of images sharing common environmental, scene, or acquisition conditions. Performance is typically summarized using a single dataset-level metric, such as mAP, which averages results across all conditions and can therefore mask substantial performance variation between domains. Although some studies have explored underwater domain generalization using synthetic domains generated through style transfer~\cite{liu2020towards, chen2023achieving}, these domains are difficult to interpret in terms of real deployment conditions. 

This missing understanding of how different underwater domains affect performance has practical consequences. Identifying domains that are consistently difficult for human annotators could help prioritize limited annotation budgets and quality-control resources. Likewise, if detectors fail systematically under particular conditions, these domains could be targeted during dataset collection or stress-testing, and inform deployment decisions, helping to identify situations where additional monitoring or model adaptation may be required. However, such recommendations require a systematic and interpretable way to characterize underwater domains, which is currently lacking.

To fill this gap, we systematically investigate, for the first time, how underwater domain conditions influence both human annotation quality and object detector performance. 
We make the following contributions:
\begin{enumerate}[itemsep=0pt, parsep=0pt, topsep=0pt]
    \item We introduce an underwater domain labeling framework to systematically characterize underwater domain shift by assigning each image a set of interpretable categorical labels that describe image appearance, scene composition, and acquisition geometry properties (Section~\ref{sec:domain-labeling}). These domain labels enable semantically meaningful grouping based on physical factors for domain-aware evaluation. 
    \item We conduct the first systematic analysis of domain-dependent annotation difficulty in underwater imagery, demonstrating that human annotation quality varies substantially with environmental conditions and providing actionable recommendations for allocating human effort and quality-control resources (Section~\ref{sec:annotation_analysis}).
    \item We conduct a large-scale analysis of domain-dependent detection performance across multiple object detection architectures, revealing consistent and sometimes counter-intuitive failure modes that are hidden by aggregate metrics (Section~\ref{sec:detection_analysis}). Our findings establish domain-aware benchmarking and stress testing as practical complements to aggregate metrics.
\end{enumerate}

\noindent The domain labels, evaluation splits, and code are available at: \href{https://github.com/melaniewille/domain-aware-benchmarking}{github.com/melaniewille/domain-aware-benchmarking}.

\section{Related Work}
\label{sec:rel_work}
Underwater object detection has received increasing attention as a challenging computer vision problem due to severe appearance variation, small and densely distributed targets, and limited annotation quality~\cite{raine2026ai}. Such variability is a major source of domain shift, while label quality remains an important concern for underwater datasets. We therefore review prior work in both areas.

\textbf{Domain Shift in Underwater Object Detection:}
Domain shift generally refers to changes in the input data distribution between training and deployment scenarios that cause performance degradation~\cite{walker2024domain}. This effect is particularly pronounced in underwater object detection because of the dynamic nature of marine environments. Water properties vary significantly across locations, seasons, weather, and depth, resulting in differences in turbidity, light scattering and absorption, contrast, and color, which in turn influence marine species and seafloor structure~\cite{walker2024domain, niu2025domain, elmezain2025advancing, wille2026all}. 

These environmental factors alter image appearance and object characteristics, causing image- and instance-level domain shift that limits detector generalization~\cite{han2023see}. Recent applied work also links performance degradation to acquisition factors such as camera movement and distance~\cite{rawlinson2025urchinbot}. However, these findings are based on manual inspection of failure cases, highlighting the lack of structured approaches to understand domain-dependent performance.

\definecolor{myblue}{RGB}{47,110,186}
\definecolor{mypurple}{RGB}{153,55,149}
\definecolor{mygreen}{RGB}{106,172,71}

\begin{figure*}
    \centering
    \includegraphics[width=\linewidth, 
        trim={0.3cm 0cm 1.4cm 0cm},
        clip
    ]{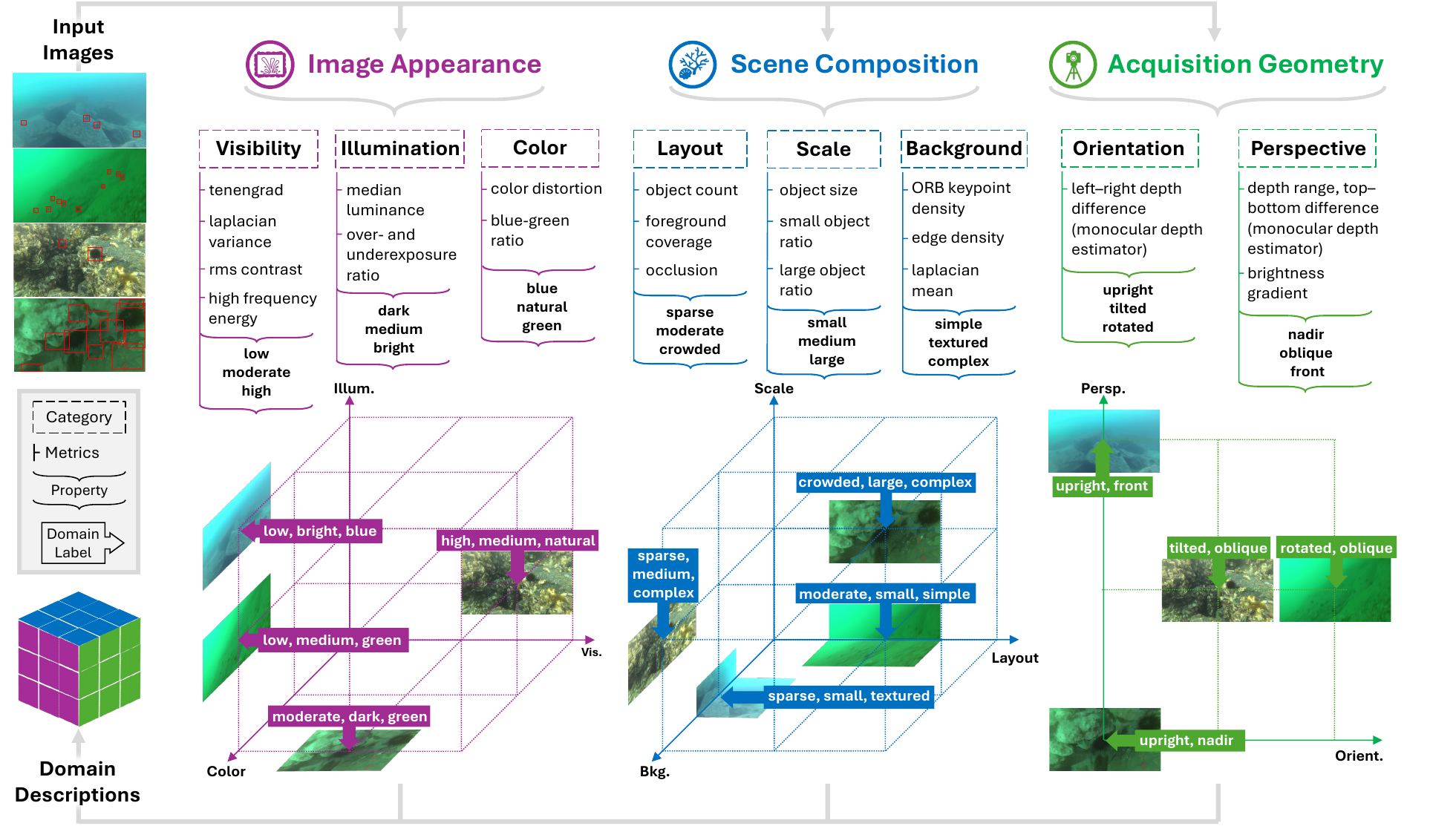}
    \vspace{-0.6cm}
    \caption{Overview of our underwater domain labeling framework. Images are characterized along three complementary axes: \textcolor{mypurple}{image appearance} (left), \textcolor{myblue}{scene composition} (middle), and \textcolor{mygreen}{acquisition geometry} (right). Each axis is decomposed into domain categories (\eg Visibility, Illumination, Color), for which selected metrics are used to assign categorical properties (\eg low, moderate, or high). The resulting properties are combined across categories into domain labels (\eg low, bright, blue) to define interpretable underwater domains.}
    \label{fig:domain_labeling_framework}
    \vspace{-0.1cm}
\end{figure*}

\textbf{Model-Centric Approaches to Domain Shift:}
Most prior works address underwater domain shift by improving detector robustness. Proposed methods include adversarial domain adaptation with enhanced images~\cite{wen2026joint}, few-shot or transfer learning adaptation~\cite{ren2026underwater, sun2025efcwm}, physics-based priors or frequency-domain representations to preserve feature information~\cite{luo2025hyuod, niu2025domain, xie2024fiod}, and architectural enhancements such as attention mechanisms and adaptive feature correction modules~\cite{saoud2024mars, sun2025efcwm}. While effective, these methods focus on learning domain-invariant representations and do not explicitly define or analyze domain-specific factors. We introduce domain labels that allow existing detectors to be stress-tested across physically meaningful underwater conditions.

\textbf{Datasets and Synthetic Domain Benchmarks:} The synthetic S-URPC2019~\cite{liu2020towards} and S-UODAC2020~\cite{chen2023achieving} datasets were proposed to study domain generalization in underwater object detection. However, they alter image appearance through synthetic style transfer which does not directly tie to intrinsic physical or semantic properties of the data and thus may not reflect the underlying environmental conditions or even assign images with different physical properties in the same synthetic domain.

\textbf{Annotation Quality in Underwater Datasets:} Noisy or inconsistent labels are a major challenge in underwater object detection~\cite{chen2024underwater,chen2025underwater,raine2026ai,lucas2025underwater}. Annotation quality depends on annotation protocols, observer expertise, and task complexity, emphasizing the need for standardization and statistical methods assessing human annotation performance~\cite{schoening2016recomia}, and differences between annotators~\cite{durden2016comparison}. Early evidence that relies on one of many existing image quality metrics~\cite{panetta2015human,wang2018imaging,yang2021reference,yang2015uciqe} suggests that annotation errors may be linked to image characteristics with image enhancement revealing large numbers of previously missed objects in RUOD~\cite{lucas2025underwater}. The subsequent release of the re-annotated RUOD-R benchmark~\cite{awad2026ruod-r} corrected these errors and demonstrated their substantial impact on detector evaluation.

Re-annotation efforts, for example in image retrieval and classification, have similarly revealed benchmark errors and label inconsistencies that can affect evaluation outcomes~\cite{radenovic2018revisiting,kisel2024flaws}, motivating methods to estimate label quality and learn with annotation noise~\cite{lad2023estimating, wei2022learning}.
However, existing work has not systematically examined whether annotation errors are domain dependent. We address this gap by comparing original RUOD and revised RUOD-R annotations under the same domain labels used for detector evaluation.

\textbf{Research Gap:} 
Existing underwater detection research has identified domain shift and annotation noise as important problems, but lacks a unified, interpretable evaluation framework for studying both. Our contribution is not a new detector or a new image-enhancement method. Instead, we make domain labels, evaluation splits, and analysis code for domain-aware benchmarking across physically meaningful and interpretable underwater domains publicly available. This enables the development and evaluation of detectors that generalize more reliably across diverse underwater environments and supports the design of targeted strategies for particularly challenging conditions.

\section{Proposed Domain Labeling Framework}
\label{sec:domain-labeling}
We propose an underwater domain labeling framework as illustrated in Fig.~\ref{fig:domain_labeling_framework}. Prior work has established that detection performance is affected by image- and instance-level factors, and by the data collection platform~\cite{han2023see, walker2024domain, rawlinson2025urchinbot}. Building on those observations, we characterize underwater domain variation using three complementary axes, that are tied to intrinsic physical and semantic image properties: image appearance, which captures image degradation caused by water properties such as turbidity, attenuation, and light scattering (Section~\ref{subsec:axis1}); scene composition, which captures the spatial arrangement of targets and the complexity of specific habitats (Section~\ref{subsec:axis2}); and acquisition geometry, which captures camera pose and viewpoint (Section~\ref{subsec:axis3}). These axes correspond to three practical sources of variability in underwater surveys: environmental conditions; habitat and target structure; and platform configuration.

Each axis is decomposed into a small set of interpretable categories to quantitatively assess the image properties using a combination of image statistics, object-level properties, and depth-based cues, which then form the domain labels. The metric definitions, thresholds, and implementation details are provided in the Supplementary Material, together with additional analyses to assess robustness, including manual sanity checks, domain co-occurrence analysis, and split-distribution comparisons.

\subsection{Axis 1: Image Appearance}
\label{subsec:axis1}
\hspace{\parindent}\textbf{Visibility.} 
Visibility describes overall clarity of an image and is affected by underwater scattering, attenuation, and blur. We characterize visibility using complementary sharpness, contrast, and frequency-based measures and categorize images as low, moderate, or high visibility.

\textbf{Illumination.}
Illumination describes global brightness and exposure conditions. Images are grouped into dark, medium, and bright categories using luminance statistics.

\textbf{Color.}
Color captures wavelength-dependent light attenuation effects that produce characteristic color casts or distortions. Based on channel imbalance measures, images are categorized as blue, green, or natural.

\subsection{Axis 2: Scene Composition}
\label{subsec:axis2}

\hspace{\parindent}\textbf{Layout.}
Layout describes object density and spatial arrangement within the image. Using object count, foreground coverage, and overlap information, images are categorized as sparse, moderate, or crowded.

\textbf{Scale.}
Scale describes the relative size of objects in the image. Based on object area statistics, images are grouped into small-, medium-, or large-object categories.

\textbf{Background.}
Background captures the amount of structural detail present in non-object regions. We quantify background complexity using texture and edge information and classify images as simple, textured, or complex.

\subsection{Axis 3: Acquisition Geometry}
\label{subsec:axis3}

\hspace{\parindent}\textbf{Orientation.}
Orientation describes the horizontal alignment of the camera to the scene. Images are categorized based on depth differences as upright, tilted, or rotated.

\textbf{Perspective.}
Perspective captures the dominant viewing direction. We rely on depth and illumination gradients to group images into nadir, oblique, or front-view categories.

\section{Domain-Dependent Annotation Difficulty}
\label{sec:annotation_analysis}
This section investigates whether annotation difficulty in underwater object detection is itself domain dependent. This is important because benchmarks usually treat ground-truth labels as uniformly reliable, yet underwater conditions may change not only detector performance but also human annotation quality. We study this using the original RUOD benchmark~\cite{fu2023ruod} and RUOD-R, its recently quality-assured and professionally re-annotated version~\cite{awad2026ruod-r}. This paired setting allows us to use annotation changes between the datasets as a proxy for domain-dependent annotation difficulty.

\begin{table}
\centering
\caption{RUOD and RUOD-R subsets that include all the images shared in both versions when filtered for our target classes. Reannotations have boosted Holothurian (Ho), Echinus (Ec), and Starfish (St) object counts by 71\%, 61\% and 56\%, respectively, and Scallop (Sc) instances even by 292\%.}
\vspace{-0.2cm}
\label{tab:ruod_study_versions}
\footnotesize
\setlength{\tabcolsep}{3pt}
\begin{tabular}{lccrrrrr}
\toprule
Datasets & \# Images & \multicolumn{5}{c}{\# Objects} \\
\cmidrule(lr){3-7}
(4-class filtered) & & Ho & Ec & Sc & St & Total \\
\midrule
RUOD      & 4,260 & 7,623  & 11,292 & 7,406  & 8,112  & 34,433 \\
RUOD-R   & 4,260 & 13,071 & 18,217 & 29,052 & 12,642 & 72,982 \\
\bottomrule
\vspace{-0.6cm}
\end{tabular}
\end{table}


\subsection{Annotation Difficulty: Method}
\label{subsec:annotation-study-method}
We compare the original and revised annotations using the 4,260 images shared by RUOD and RUOD-R after filtering to the four common target classes: echinus, holothurian, scallop, and starfish (Table~\ref{tab:ruod_study_versions}). Revision has increased instance counts across all species, resulting in more than double the original total objects, indicating that the initial RUOD annotations primarily missed (false negative) objects rather than over-counted (false positive). Domain labels are assigned using the proposed framework (Section~\ref{sec:domain-labeling}) based on the revised RUOD-R annotations, which are assumed to more accurately represent the true scene composition. Using the RUOD annotations would propagate their known incompleteness into the domain assignments, which could result in misclassification of images with small unannotated objects into the sparse layout or the large-object scale categories.

To quantify annotation difficulty, we compute an image-level correction rate $C_i = \frac{A_i + R_i}{O_i}$, where $A_i$, $R_i$, and $O_i$ denote the number of added\footnote{Revised annotations are considered newly added if they do not match an original annotation at IoU threshold 0.5.}, removed\footnote{Original annotations are considered removed if they do not match a revised annotation at IoU threshold 0.5.}, and original\footnote{All images in the original dataset were verified to contain at least one annotation (${O_i} \neq 0$); therefore, rates are well-defined for every image.} annotations in image $i$, respectively. Higher values indicate greater disagreement between annotation versions. 

We analyze correction rates using their mean and median values for each domain category. Since the correction rates are skewed and non-normal (see Supplementary Material), we use non-parametric statistical tests: we first apply a Kruskal-Wallis test to assess whether correction rates differ across categories and report epsilon-squared $\varepsilon^2$ as an effect-size measure. We then perform pairwise Mann--Whitney U tests with Benjamini--Hochberg correction to identify which category pairs differ significantly. For ordinal properties, we also report Spearman's rank correlation coefficients $\rho$ to quantify the strength and direction of monotonic trends. Finally, we perform bootstrap resampling with 1,000 iterations to assess whether the observed effects are stable.


\begin{figure}
    \centering
    \includegraphics[
        width=\linewidth,
        trim={0.2cm 0cm 0cm 0cm},
        clip
    ]{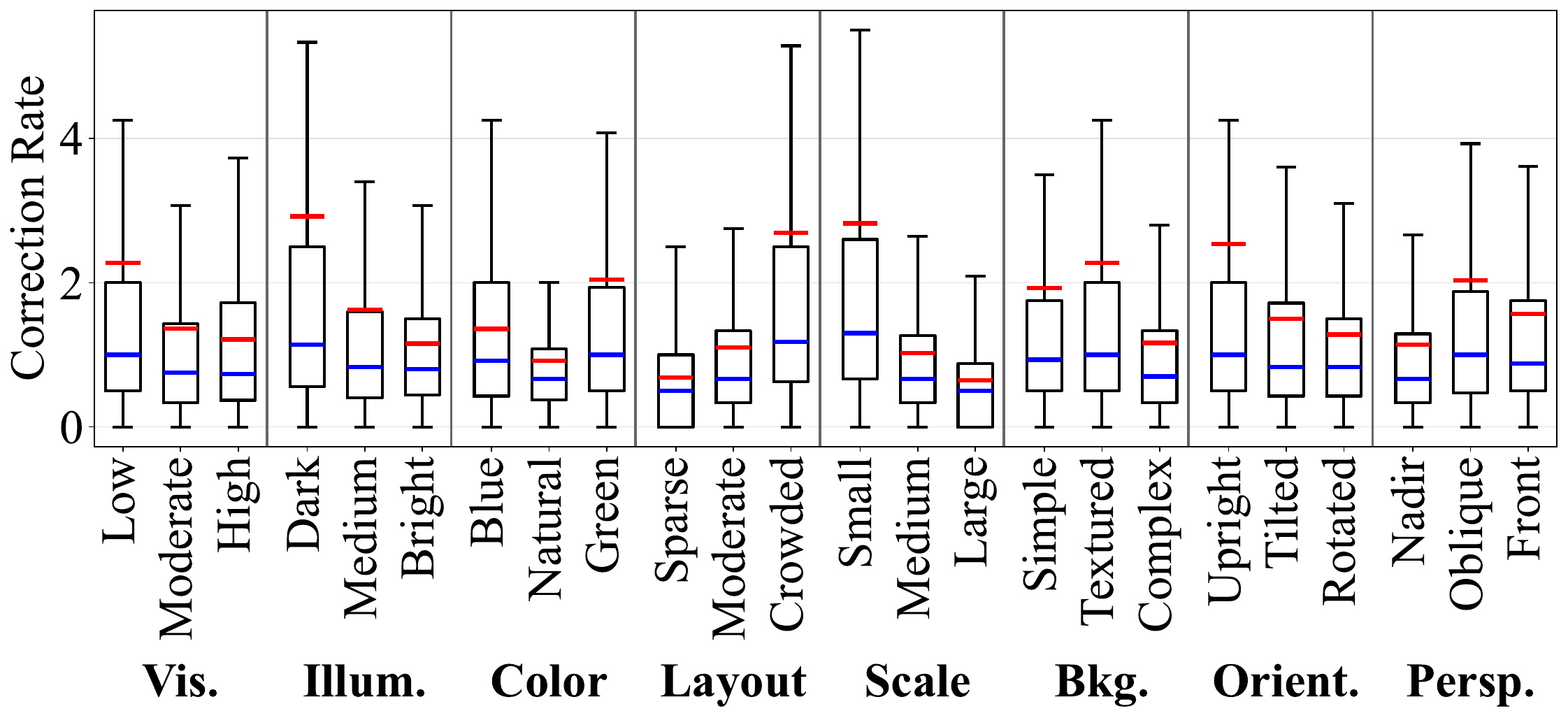}
    \vspace{-0.6cm}
    \caption{Boxplots of correction rates grouped by domain category. Red and blue lines denote the mean ($\mu_C$) and median correction rates, respectively. The distributions reveal considerable variability in annotation disagreement across domains.}
    \label{fig:mean-median-correction-rate}
    \vspace{-0.4cm}
\end{figure}

\subsection{Annotation Difficulty: Results}
\label{subsec:annotation-study-results}

Fig.~\ref{fig:mean-median-correction-rate} illustrates the correction rates across domain categories. There are substantial differences in both central tendency and spread of the distributions. Means consistently exceed medians, indicating strongly right-skewed distributions, where most images require few corrections while a small number contain many annotation errors. To disentangle the connection between image properties and annotation errors, we look at every domain category individually. Kruskal-Wallis tests in Table~\ref{tab:statistical-significance} show that all investigated domain properties significantly affect annotation quality (all p-values $<0.001$), but the corresponding effect sizes differ substantially, as described in the following.

\subsubsection{Scene Composition As Primary Driver}
Scale and layout show the strongest effects on annotation difficulty across all categories. Small objects ($\mu_C{_\text{small}}\!=\!2.8$) and crowded scenes ($\mu_C{_\text{crowded}}\!=\!2.7$) have correction rates around four times higher than large objects and sparse scenes. This suggests that limited object size, dense aggregations, overlap, and ambiguous boundaries increase the likelihood of missing instances.

The statistical analysis supports this interpretation: Kruskal-Wallis tests identify scale ($\varepsilon^2{_\text{scale}}\!=\!0.137$) and layout ($\varepsilon^2{_\text{layout}}\!=\!0.115$) as the dominant drivers of annotation difficulty, and all pairwise comparisons between their category properties are significant. Spearman correlations further reveal strong monotonic relationships for layout ($\rho{_\text{layout}}\!=\!0.34$) and scale ($\rho{_\text{scale}}\!=\!-0.37$): annotation difficulty consistently increases from sparse to crowded scenes and decreases from small to large objects.

Background complexity shows a weaker and non-monotonic relationship. Textured backgrounds have the highest correction rates, but the effect is smaller compared to the above ($\varepsilon^2{_\text{bkg}}\!=\!0.026$). Since the background category is derived from low-level structural information, it may be less relevant to human annotators that likely rely more heavily on semantic context and object-background contrast.

\begin{table}
\centering
\footnotesize
\caption{Statistical analysis of annotation correction rates across domain properties. All Kruskal--Wallis tests were significant ($p<0.001$), therefore only epsilon-squared effect sizes are reported. Pairwise comparisons were performed using Mann--Whitney U tests with Benjamini--Hochberg correction, and only statistically significant category pairs are listed. Spearman correlations are reported for ordered categories with monotonic trends.}
\vspace{-0.2cm}
\label{tab:statistical-significance}
\setlength{\tabcolsep}{3pt}
\begin{tabular}{lccc}
\toprule
Category & Kruskal $\varepsilon^2$ & Significant pairs & Spearman $\rho$ \\
\midrule
Visibility   & 0.020 & low--moderate, low--high & -0.13\ \\
Illumination & 0.019 & dark--medium, dark--bright & -0.12 \\
Color        & 0.013 & natural--blue, natural--green & -- \\
Layout       & 0.115 & all pairs & 0.34 \\
Scale        & 0.137 & all pairs & -0.37 \\
Background   & 0.026 & all pairs & -- \\
Orientation  & 0.004 & upright--tilted, upright--rotated & -0.06 \\
Perspective  & 0.009 & nadir--oblique, front--nadir & -- \\
\bottomrule
\vspace{-0.6cm}
\end{tabular}
\end{table}

\subsubsection{Image Appearance Has Secondary Influence}
Image appearance also affects annotation discrepancies, but with smaller effect sizes than scene composition. Dark images ($\mu_C{_\text{dark}}\!=\!2.9$) have correction rates about 2.5 times higher than bright images ($\mu_C{_\text{bright}}\!=\!1.1$), and low-visibility images ($\mu_C{_\text{low}}\!=\!2.3$) have correction rates about 1.9 times higher than high-visibility images ($\mu_C{_\text{high}}\!=\!1.2$). Humans show best annotation performance on natural-color images ($\mu_C{_\text{natural}}\!=\!0.9$), whereas green images have more than double the correction rate ($\mu_C{_\text{green}}\!=\!2.0$). These trends are consistent with the intuition that reduced contrast, scattering, and color distortion make targets harder to identify.

However, compared to layout and scale, the Kruskal effect sizes are smaller ($\varepsilon^2_{\text{vis}}\!=\!0.020$, $\varepsilon^2_{\text{illum}}\!=\!0.019$)  and monotonic correlations weaker ($\rho_{\text{vis}}\!=\!-0.13$, $\rho_{\text{illum}}\!=\!-0.12$). Pairwise comparisons reveal that annotation difficulty increases primarily under severe image degradations: moderate and high visibility levels are statistically similar, whereas low-visibility and dark conditions show significantly higher correction rates (Table~\ref{tab:statistical-significance}). Likewise, both blue- and green-distorted images are significantly more difficult to annotate than natural images.

\subsubsection{Acquisition Geometry Plays Minor Role}
Acquisition geometry shows the weakest association with annotation difficulty. Although upright images ($\mu_C{_\text{upright}}\!=\!2.5$) and oblique views ($\mu_C{_\text{oblique}}\!=\!2.0$) unexpectedly show higher average correction rates than other geometric categories, the associated effect sizes are negligible ($\varepsilon^2_{\text{orient}}\!=\!0.004$, $\varepsilon^2_{\text{persp}}\!=\!0.009$), suggesting that these differences are likely caused by interactions with other domain factors rather than geometry itself.

\subsection{Summary of Annotation Difficulty}
Overall, annotation discrepancies between RUOD and RUOD-R are strongly domain dependent. The largest effects are associated with scene composition, especially object scale and layout. Image appearance in terms of the visual characteristics perceived by the annotator has a secondary but consistent influence. In contrast, acquisition geometry contributes little to annotation difficulty. Bootstrap analysis confirms that these trends remain stable across repeated resampling, while an IoU threshold sensitivity analysis shows that the observed domain gaps persist across different matching thresholds (see Supplementary Materials). These results indicate that annotation reliability should not be treated as uniform across underwater benchmarks. Instead, domain-aware evaluation is needed not only for detector performance, but also for auditing label quality and identifying images that require greater annotation effort or quality control.

\section{Domain-Dependent Detector Performance}
\label{sec:detection_analysis}
We next ask whether the proposed domain labels from Section~\ref{sec:domain-labeling} reveal systematic detector performance gaps. 
By analyzing whether different object detector architectures fail under similar underwater domain conditions, we provide a direct test of the proposed labels as an evaluation tool for stress-testing object detectors.

\begin{figure}
    \centering
    \includegraphics[width=\linewidth, trim=0.5cm 0cm 0cm 0.2cm, clip]{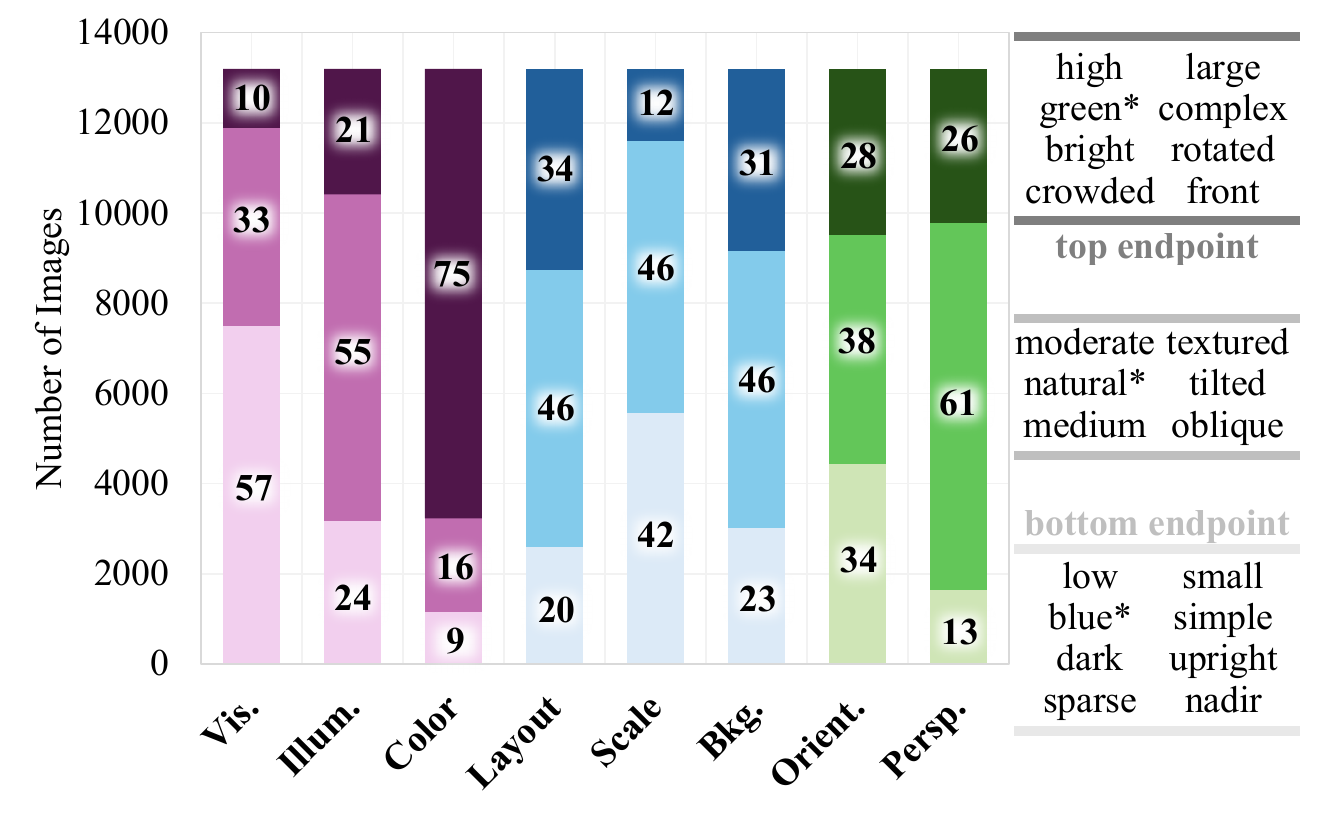}
    \vspace{-0.7cm}
    \caption{Distribution of domain properties in the 13,196 training images. Stacked bar segments represent properties in each category, with numbers showing relative proportions (\%). Evaluation is performed only on bottom and top endpoints for all categories except color, which has non-ordered properties (indicated by *).}
    \label{fig:train_domain_distr}
    \vspace{-0.25cm}
\end{figure}

\subsection{Detector Performance: Method}
\label{subsec:detection-study-method}
We combine data from three widely used underwater object detection datasets: Detecting Underwater Objects (DUO)~\cite{liu2021duo}, Real-world Underwater Object Detection, in its new re-annotated version (RUOD-R)~\cite{awad2026ruod-r}, and Underwater Target Detection Algorithm Competition 2020 (UTDAC)~\cite{dai2024erl-net}. Since echinus, holothurian, scallop, and starfish are the only classes shared in these datasets, they form our target classes. Domain labels are assigned using the proposed framework (Section~\ref{sec:domain-labeling}). 
Fig.~\ref{fig:train_domain_distr} shows the resulting domain distribution.

The combined dataset contains 18,852 images and 196,102 annotations. We split the data into training (70\%), validation (10\%), and test (20\%) subsets. 
We train three representative detector architectures: YOLO26n (one-stage, anchor-free)~\cite{jocher2026ultralytics}, Faster RCNN (two-stage, anchor-based)~\cite{ren2016faster}, and RT-DETR (transformer)~\cite{zhao2024detrs}. All models are trained on the full mixed-domain training set for 50 epochs using their respective default hyperparameters, and the best validation checkpoint is used for evaluation. We then evaluate each model separately on the domain categories within the test split. We verify that all observed trends are stable by repeating the experiments five times with different randomized train/validation/test splits.

Performance is reported as the mean and standard deviation across the five independent runs (see Supplementary Material), using standard mAP50 and mAP50--95. For ordered categories, the main paper focuses on the two endpoint properties to quantify performance differences between the most distinct domain conditions. For color, which is non-ordered, we report all categories.


\subsection{Detector Performance: Results}
\label{subsec:detection-study-results}

\begin{figure}
    \centering
    \includegraphics[
        width=\linewidth,
        trim={0.25cm 0.25cm 0.1cm 0cm},
        clip
    ]{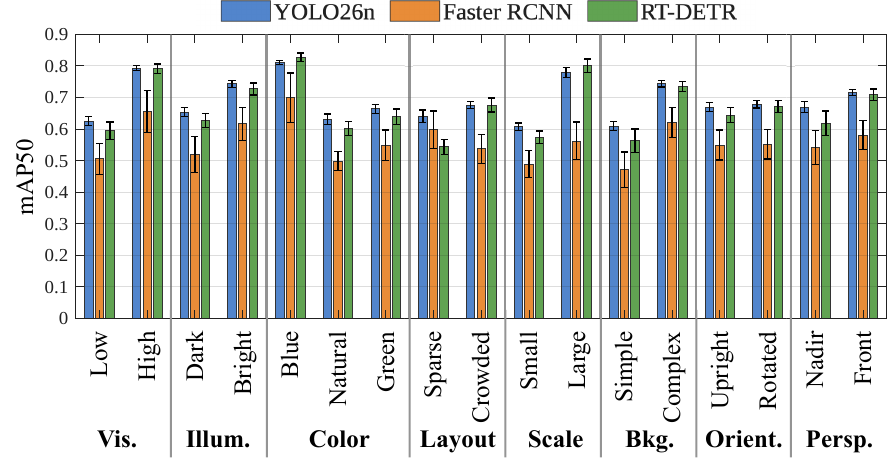}
    \vspace{-0.6cm}
    \caption{mAP50 per domain and architecture, averaged over five runs. Error bars indicate standard deviation. YOLO26n (blue), Faster RCNN (orange) and RT-DETR (green) show broadly consistent performance trends across domain properties, except for layout, which exhibits architecture-dependent differences.}
    \label{fig:mAP-across-domains}
    \vspace{-0.3cm}
\end{figure}

Fig.~\ref{fig:mAP-across-domains} shows detection performance across domain categories. Performance varies substantially across axes, and several gaps are consistent across the three detector architectures (Table~\ref{tab:map50_gaps}). This indicates that the proposed domain labels capture systematic domain-dependent differences that are not visible from aggregate mAP alone.

\subsubsection{Challenging Domain Properties}
Several domain properties show performance trends consistent with standard detection intuition. Low-visibility images consistently underperform compared to high-visibility images, with mAP50 drops of 17 points on average (Table~\ref{tab:map50_gaps}). This is expected, as blur, haze, and reduced contrast weaken object boundaries and remove fine visual details. Visibility exhibits one of the largest performance gaps, making it a dominant environmental factor.

Illumination shows a weaker but consistent association with detector performance. Bright scenes outperform dark scenes across all detectors, with a 10-point gap in mAP50. The consistency across detectors suggests that reduced signal quality in dark images represents a general challenge rather than an architecture-specific weakness.

Scale also exhibits a strong effect, consistent with the well-established difficulty of small object detection. Small objects consistently perform the worst across all architectures, although the magnitude varies, with the largest mAP50 gap of 23 points for RT-DETR. This observation agrees with the limited visual information available for small targets.

Perspective is associated with moderate performance differences: front-view images consistently outperform nadir views by $\approx$6 points on average. Species-level analysis (Supplementary Material) reveals that this difference is particularly pronounced for scallops. A possible explanation is that viewed from above, scallops often appear as relatively flat and partially buried, blending into the seabed, whereas front views reveal more distinctive shell features.


\newcommand{\gapcell}[1]{%
  \cellcolor{red!#1}%
}

\begin{table}
\centering
\caption{mAP50 gaps across architectures. Similar values suggest effects are largely detector-independent, except for layout, where Faster RCNN shows an opposing trend to YOLO26n and RT-DETR.}
\vspace{-0.2cm}
\label{tab:map50_gaps}
\scriptsize
\setlength{\tabcolsep}{2.3pt}
\begin{tabular}{lcccccccc}
\toprule
 & Vis. & Illum. & Color & Layout & Scale & Bkg. & Orient. & Persp. \\
 \midrule
 & high & bright & blue & crowded & large & complex & rotated & front \\
  & -low & -dark & -nat & -sparse & -small & -simple & -upright & -nadir \\
\midrule
YOLO26n
& \cellcolor{red!28}0.17
& \cellcolor{red!16}0.09
& \cellcolor{red!29}0.18
& \cellcolor{red!8}0.04
& \cellcolor{red!28}0.17
& \cellcolor{red!19}0.13
& \cellcolor{red!1}0.01
& \cellcolor{red!10}0.05 \\

Faster RCNN
& \cellcolor{red!22}0.15
& \cellcolor{red!17}0.10
& \cellcolor{red!33}0.20
& \cellcolor{red!12}-0.06
& \cellcolor{red!14}0.07
& \cellcolor{red!22}0.15
& \cellcolor{red!0}0.00
& \cellcolor{red!8}0.04 \\

RT-DETR
& \cellcolor{red!33}0.20
& \cellcolor{red!17}0.10
& \cellcolor{red!38}0.23
& \cellcolor{red!19}0.13
& \cellcolor{red!38}0.23
& \cellcolor{red!28}0.17
& \cellcolor{red!1}0.01
& \cellcolor{red!12}0.06 \\

\midrule
\textbf{Mean $|$Gap$|$}
& \cellcolor{red!28}\textbf{0.17}
& \cellcolor{red!17}\textbf{0.10}
& \cellcolor{red!33}\textbf{0.20}
& \cellcolor{red!15}\textbf{0.08}
& \cellcolor{red!24}\textbf{0.16}
& \cellcolor{red!22}\textbf{0.15}
& \cellcolor{red!1}\textbf{0.01}
& \cellcolor{red!12}\textbf{0.06} \\
\bottomrule
\vspace{-0.5cm}
\end{tabular}
\end{table}

\subsubsection{Counter-Intuitive Performance Patterns}
Other domain properties show trends that are less obvious from standard assumptions and difficult to identify from aggregate metrics alone. The color category reveals an interesting mismatch between detection performance and training data quantity: Blue images consistently achieve the highest performance despite being underrepresented in training data (Fig.~\ref{fig:train_domain_distr}): The mAP gap reaches more than 20 points between blue and natural images. This suggests that color labels capture more than simple sample frequency and may reflect correlated environmental conditions. For example, green water is often associated with phytoplankton and suspended particles,~\ie more scattering and less contrast, while blue water is often found in clearer, better illuminated sites. Consistent with this interpretation, blue images are disproportionately associated with high visibility (60.1\% \vs 9.8\% overall) and brightness (47.9\% \vs 21.0\% overall). This suggests that domain properties interact.

Unlike other domain categories, layout performance trends vary across architectures. 
RT-DETR achieves its highest performance in crowded scenes, outperforming sparse scenes by 13 percentage points. In contrast, YOLO26n and Faster R-CNN perform best in moderate layouts. However, YOLO26n still performs better in crowded than sparse scenes (4 points), whereas Faster R-CNN shows the opposite trend (6 points). This suggests that architectural differences may play a role. Faster RCNN uses Non-Maximum Suppression (NMS), which may delete detections of crowded objects with highly overlapping bounding boxes, whereas YOLO26n and RT-DETR are both NMS-free. 





The background category also follows a counter-intuitive pattern: complex scenes substantially outperform simple backgrounds across all three detectors, with performance gaps of 13--17 mAP50 points. Co-occurrence analysis suggests that this effect is closely linked to visibility. Almost all simple-background images occur in low-visibility conditions, whereas complex backgrounds are predominantly associated with moderate and high visibility. Since reduced visibility suppresses texture and structural detail, causing backgrounds to appear simpler, the observed performance differences are likely driven primarily by image quality.

Orientation has the weakest effect on performance. Rotated images slightly outperform upright ones, but the mAP50 gap of $\approx$1 point remains negligible across architectures. This suggests that modern detectors are relatively robust to moderate changes in object orientation.

\subsection{Summary of Detector Performance}

Overall, the results reveal that detection performance is influenced notably by all domain categories except orientation. Substantial gaps between category extremes indicate clearly favorable and unfavorable conditions for a detector. Several counter-intuitive trends further suggest that interactions between domain properties and scene context play an important role. Small variation of mAP50 values across runs (Fig.~\ref{fig:mAP-across-domains}) and largely consistent trends across architectures (Table~\ref{tab:map50_gaps}) suggest that the observed performance differences reflect characteristics of the data. These results demonstrate the value of domain-aware evaluation for identifying detector weaknesses that are hidden by aggregate mAP.

\section{Discussion}
\label{sec:discussion}

\noindent\textbf{Task-Dependent Sources of Difficulty:}
\label{subsec:image-difficulty}
Table~\ref{tab:effect_comparison} summarizes how different domain factors affect human annotators and object detection models. Underwater image difficulty is not universal, but depends on both the task,~\ie annotation or detection, and interactions between domain properties.

Scale, visibility, and to a lesser degree illumination consistently influence both human annotators and object detector performance similarly. Larger objects, higher visibility and more light make it significantly easier to identify marine targets in an underwater image for humans as well as detectors. In contrast, changes in camera orientation are comparatively less important to either, while a front-facing perspective might support the detector for certain species. 

Habitat structure affects humans and detectors differently. Textured backgrounds and crowded layouts increase annotation difficulty, likely due to overlap and visual clutter, yet detectors consistently perform better in complex scenes and RT-DETR and YOLO26n better in crowded layout. Some detector architectures may handle crowded scenes well, whereas human observers generally face challenges from visual clutter and annotation overload. Color shows another mismatch: humans annotate natural images most reliably, whereas the best detection performance was recorded in blue-distorted images, which suggests that color itself is less significant to detectors than correlated image properties.

These findings indicate that difficult domains should be interpreted differently for annotation and detection. Images with small targets in low-visibility, dark conditions are challenging for both, but structurally complex scenes may be favorable for detectors while still requiring additional annotation quality control. Domain-aware evaluation is useful because it separates task-dependent failure patterns, rather than treating dataset difficulty as a single global property.

\begin{table}
\centering
\caption{Qualitative comparison of domain-associated difficulty for human annotation and detector performance. Difficulty levels are derived from the relative effect sizes observed in Sections~\ref{sec:annotation_analysis} and~\ref{sec:detection_analysis}. Agreement indicates whether humans and detectors exhibit similar responses to a domain property ($\checkmark$) or not ($\times$),~\eg complex backgrounds challenged humans but benefited detectors, hence $\times$.}
\vspace{-0.2cm}
\label{tab:effect_comparison}
\scriptsize
\setlength{\tabcolsep}{0.8pt}
\renewcommand{\arraystretch}{1.15}
\begin{tabular}{lcccccccc}
\toprule
& Vis. & Illum. & Color & Layout & Scale & Bkg. & Orient. & Persp. \\
\midrule
Annotator
& moderate
& moderate
& moderate
& strong
& strong
& moderate
& weak
& weak \\
Detector
& strong
& moderate
& strong
& moderate
& strong
& strong
& weak
& moderate \\
\midrule
Agreement
& $\checkmark$
& $\checkmark$
& $\times$
& $\times$
& $\checkmark$
& $\times$
& $\checkmark$
& $\checkmark$ \\
\bottomrule
\vspace{-0.55cm}
\end{tabular}
\end{table}

\vspace{0.15cm}
\textbf{Implications for Dataset Design:}
\label{subsec:dataset-design}
Physically meaningful domain labels provide a practical basis for improving and auditing annotation reliability, dataset construction and training. We recommend:

\textit{(i) Allocate annotation budget according to domain difficulty.}
Since annotation difficulty is not uniformly distributed across underwater data, difficult domains may require disproportionately more annotation effort.

\textit{(ii) Prioritize preprocessing for challenging domains.}
Preprocessing techniques such as contrast and illumination enhancement or color correction can be selectively applied to difficult images before labeling to assist annotators.

\textit{(iii) Incorporate domain information into annotation quality control.}
Domains highly prone to annotation noise may benefit from repeated verification and expert review.

\textit{(iv) Use domain labels to guide data collection and balancing.}
Domain labels can identify underrepresented or challenging conditions, so that future dataset construction can target specific domains or adopt strategies that reduce avoidable sources of difficulty,~\eg improved lighting.

\textit{(v) Incorporate domain information into noise-aware learning frameworks.}
Domain labels may provide useful prior information for training algorithms that account for label noise,~\eg by assigning lower confidence.

\vspace{0.15cm}
\textbf{Implications for Detector Evaluation:}
\label{subsec:detector-evaluation}
Aggregate metrics such as dataset-level mAP are insufficient for characterizing underwater detector robustness. Physically meaningful domain labels provide a foundation for more informative and systematic evaluation, as well as future benchmark and model development. We recommend:

\textit{(i) Report domain-wise performance alongside aggregate metrics.}
Similar to class-wise evaluation, domain-wise analysis offers an additional dimension for understanding model behavior and identifying failure modes.

\textit{(ii) Stress-test detectors on challenging domain properties.}
Performance should be assessed under worst-case conditions identified by the analysis to quantify robustness.

\textit{(iii) Report domain-specific performance gaps.}
Quantifying the performance difference between favorable and challenging domain properties measures sensitivity.

\textit{(iv) Use domain labels to construct domain-shift benchmarks.}
Domain labels enable benchmark splits that isolate specific underwater conditions for systematic analyses.

\textit{(v) Use domain analysis to guide algorithm development.}
Identified failure modes can direct future research toward the conditions where current detectors struggle most.

\vspace{0.15cm}
\textbf{Limitations and Future Work:}
\label{subsec:limitations}
First, since many underwater characteristics are naturally correlated, it is difficult to attribute observed effects to single domain properties. Therefore, further investigations of the interactions and higher-dimensional domain descriptions would be useful to better understand their relative contributions. 

Second, our proposed labels discretize continuous image, scene, and geometry characteristics into categorical properties. This improves interpretability for domain-wise analysis, but different metrics or thresholds may change domain assignments. Future work may therefore explore continuous or learned representations of underwater domains. 

Third, scene composition labels depend on object annotations and, in the annotation study, are derived from the revised RUOD-R annotations to better reflect the true scene. However, this dependence may partly contribute to the observed associations with correction rates and limits the framework's applicability to previously unannotated datasets. Future work may explore annotation-independent estimates of scene composition.

Fourth, our study focuses on four benthic target classes. Extending the framework to additional underwater datasets, species, and related tasks such as instance segmentation would provide further insight into the generality.

Finally, the proposed domain labels provide a useful basis for developing domain-specific augmentation, adaptation, and training strategies to improve robustness in challenging underwater environments.

\section{Conclusion}
\label{sec:conclusion}

We studied underwater domain shift from a data-centric perspective and proposed a framework for assigning physically meaningful domain labels to analyze domain-dependent annotation difficulty and detector performance. By exposing systematic performance differences in both, our framework transforms domain shift from an uncontrolled, hidden source of benchmark variability into a measurable evaluation dimension with implications for how underwater datasets are constructed and evaluated, shaping the next generation of underwater benchmarks. \\

\iftoggle{wacvfinal}{
    {
\small
\textbf{Acknowledgments}
\label{sec:acknowledgements}
This research was supported by the QUT Centre for Robotics, QUT Digital Research Infrastructure team for HPC, and an ARC DECRA Fellowship DE240100149 to TF.
}

}{}

\makeatletter
\let\wacv@origbibitem\bibitem
\def\bibitem{\ifnum\value{NAT@ctr}=32 \vfill\newpage\fi\wacv@origbibitem}
\makeatother

\small
\bibliographystyle{ieeenat_fullname}
\bibliography{main}

\end{document}